# AGIL: Learning Attention from Human for Visuomotor Tasks

Ruohan Zhang[1*], Zhuode Liu[2], Luxin Zhang[3], Jake A. Whritner[4],
Karl S. Muller[4], Mary M. Hayhoe[4], Dana H. Ballard[1]

[1]Department of Computer Science, University of Texas at Austin
[2]Google Inc.
[3]Department of Machine Intelligence, Peking University
[4]Center for Perceptual Systems, University of Texas at Austin
*zharu@utexas.edu

**Abstract**

When intelligent agents learn visuomotor behaviors from human demonstrations, they may benefit from knowing where the human is allocating visual attention, which can be inferred from their gaze. A wealth of information regarding intelligent decision making is conveyed by human gaze allocation; hence, exploiting such information has the potential to improve the agents' performance. With this motivation, we propose the AGIL (Attention Guided Imitation Learning) framework. We collect high-quality human action and gaze data while playing Atari games in a carefully controlled experimental setting. Using these data, we first train a deep neural network that can predict human gaze positions and visual attention with high accuracy (the gaze network) and then train another network to predict human actions (the policy network). Incorporating the learned attention model from the gaze network into the policy network significantly improves the action prediction accuracy and task performance.
**Keywords**: Visual Attention; Eye Tracking; Imitation Learning

## 1 Introduction

In end-to-end learning of visuomotor behaviors, algorithms such as imitation learning, reinforcement learning (RL), or a combination of both, have achieved remarkable successes in video games [27], board games [36, 37], and robot manipulation tasks [23, 29]. One major issue of using RL alone is its sample efficiency, hence in practice human demonstration can be used to speedup learning [36, 5, 14].

Imitation learning, or learning from demonstration, follows a student-teacher paradigm, in which a learning agent learns from the demonstration of human teachers [1]. A popular approach is behavior cloning, i.e., training an agent to predict (imitate) demonstrated behaviors with supervised learning methods. Imitation learning research mainly focuses on the student–advancing our understanding of the learning agent–while very little effort is made to understand the human teacher. In this work, we argue that understanding and modeling the human teacher is also an important research issue in this paradigm. Specifically, in visuomotor learning tasks, a key component of human intelligence–the visual attention mechanism–encodes a wealth of information that can be exploited by a learning algorithm. Modeling human visual attention and guiding the learning agent with a learned attention model could lead to significant improvement in task performance.

We propose the *Attention Guided Imitation Learning (AGIL)* framework, in which a learning agent first learns a visual attention model from human gaze data, then learns how to perform the visuomotor task from human decision data. The motivation is that for deep imitation learning tasks where the decision state is often in raw pixel space, the introduction of attention could help resolve two issues:

1. Humans have a unique sensory system that is different from machines' and this leads to different perceived decision states.

2. The traits of this sensory system lead to gaze behaviors and visual attention–intelligent mechanisms that are not yet available to the learning agent. Without these mechanisms, it is difficult for the agent to infer which visual features are being attended and are relevant for the decision at a given moment in a high dimensional feature space.

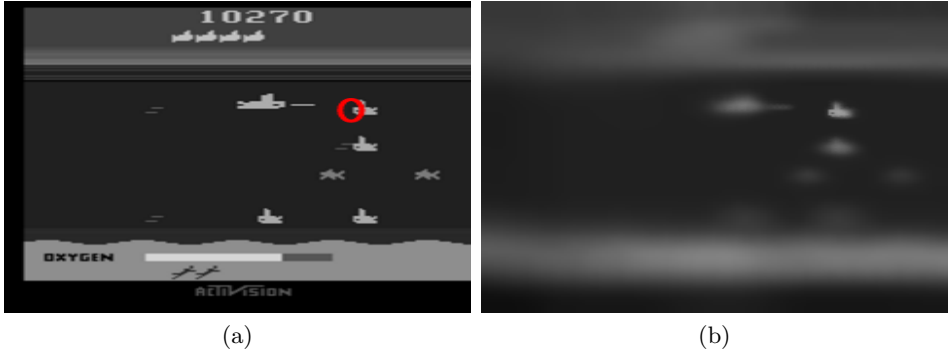

(a)                      (b)

Figure 1: An original game frame for Atari Seaquest with a red circle indicating the gaze position (a). The gaze position is used to generate foveated images (b) that are biologically plausible retinal representations of the visual stimulus (the stimulus as perceived by the human).

To elaborate the first point, humans have high acuity foveal vision in the central 1-2 visual degrees of the visual field (i.e., covering the width of a finger at arms length), with resolution decreasing in the periphery. This leads to a discrepancy in the perceived states of a human and a machine, where the machine perceives full resolution images like in Fig. 1a while a human would see Figs. 1b if the stimulus is $64.6 \times 40.0$cm and the distance to the subjects' eyes is 78.7cm. A foveal visual system may seem inferior compared to a full resolution camera, but it leads to an outstanding property of human intelligence: Visual attention, which can be seen as a feature selection mechanism. Humans manage to move their foveae to the correct place at the right time in order to emphasize important task-relevant features [7, 35]. In this way, a wealth of information is encoded in human gaze behaviors–for example, the priority of one object over another in performing an action.

Given the rich information encoded in human gaze, we hypothesize that a promising approach to strengthen an imitation learning algorithm is to model human visual attention through gaze, and subsequently include such a model in the decision-learning process. Doing so would allow the learning agent to use gaze information to help decipher the internal state representation used by the human teacher. By extracting features that are most important for tasks, the learning agent can better understand and imitate a human teachers' behaviors.

We start by extracting the large amount of high-quality training data necessary for training. With modern high-speed eye trackers we collect human game playing data and gaze data for various Atari games. We first train a deep neural network that can predict human gaze positions and visual attention (the gaze network). Second, we train another deep neural network–guided by visual attention information–to predict human actions (the policy network). Finally, we evaluate the imitation learning results in terms of both behavior matching accuracy and game playing performance.

## 2 Related Work

### 2.0.1 Modeling visual attention: bottom-up vs. top-down

Previous work in computer vision has formalized visual attention modeling as a saliency prediction problem where saliency is derived from image statistics, such as intensity, color, and orientation in the classic Itti-Koch model [16]. In recent years, this approach has made tremendous progress due to large benchmark datasets such as [20, 18] and deep neural networks [4, 24, 21, 41]. Many of saliency datasets collect human eye tracking data in a "free-viewing" manner due to their task-free nature [18].

In contrast to this approach, top-down models emphasizes the effects of task-dependent variables on visual attention [11, 38, 32]. [35] have shown that varying task instructions drastically alters the gaze distributions on different categories of objects (e.g., task-irrelevant objects are ignored even though they are salient from a bottom-up perspective). [19] has shown that in an urban driving environment, attention shifts can be accurately predicted by changing both the relative reward of different tasks and the level of uncertainty in the state estimation. The top-down attention model is hence closely related to reinforcement learning since they both concern visual state features that matter the most for acquiring the reward [11, 22].

Regardless of their approaches, these works argue that there is much valuable information encoded in gaze behaviors. It should be said that the two approaches are not mutually exclusive, since attention is modulated in both saliency-driven and volition-controlled manners [16]. As mentioned before, deep neural networks have been a standard approach to predict bottom-up saliency. In contrast, top-down gaze models often rely on manually defined task variables. Our approach seeks to combine these approaches and use the representation learning power of deep networks to extract task-relevant visual features, given task-driven gaze data. A recent work that also takes this approach is [30], where they predict human gaze while driving from raw images using a multi-channel deep network.

### 2.0.2 Attention in visuomotor learning tasks

While visual attention models have shown very promising results in several visual learning tasks, including visual question answering [42], image generation [10], image caption generation [40], and spatial transformer networks [17], incorporating visual attention models into visuomotor learning is yet to be explored.

The relation between attention and reinforcement learning has been revealed by neuroscience researchers [8, 34, 22, 15]. [28] attempts to jointly learn attention and control, and show that a learned attention model can predict visual attention much better than bottom-up saliency models. [39] show that different network components learn to attend to different visual features, but they do not explicitly model visual attention. [26] pioneered efforts to combine deep reinforcement learning and visual attention, where attention is treated as a sequential decision problem (where to look) and is jointly trained with the control policy (what to do) via deep reinforcement learning. Therefore their attention model is a non-differentiable (or "hard") attention model which leads to a computationally expensive training procedure. In contrast, saliency approaches in general prefer a differentiable (or "soft") attention model that could be trained more efficiently. Our work treats visual attention as an auxiliary component for visuomotor learning tasks and chooses to use the more efficient soft attention model.

## 3  Data Acquisition

We collected human game-playing actions using Atari games in the Arcade Learning Environment [2]. At each time step $t$, the raw image frame $I_t$, the human keystroke action $a_t$, and the gaze position $g_t$ were recorded. The gaze data was recorded using an EyeLink 1000 eye tracker at 1000Hz. The game screen is $64.6 \times 40.0$cm and the distance to the subjects' eyes was 78.7cm. The visual angle of an object is a measure of the size of the object's image on the retina. The visual angle of the screen was $44.6 \times 28.5$ visual degrees. Standard eye tracking calibration and validation techniques were implemented, resulting in an average gaze positional error of 0.44 visual degrees (covering the half width of a finger at arms length).

Our goal is to obtain the best possible control policy from human subjects, hence we take into account the limitations of human reaction time and fatigue. For visuomotor tasks like playing Atari games, human response time to visual stimulus is on average 250ms, so running the game at the original speed is too challenging for most human subjects. To allow for enough response time, the games only proceed when the subject makes an action (presses a key or keeps a key pressed down). To reduce fatigue, players play for 15 minutes and rest for at least 20 minutes.

We carefully chose eight Atari games from [27] that each represent a different genre. The data are from three amateur human players that contains a total length of 1,335 minutes and 1,576,843 image frames. The frames without gaze–due to blinking, off-screen gaze, or tracker error–are marked as bad data and excluded from training and testing (5.04%). Data belong to a single trajectory (an episode in Atari game) will not be split into training and testing set. Due to the high sampling frequency of the data recording device, two adjacent frames/actions/gazes are highly similar so we avoid putting one in training and another in testing. The sample size information for training and testing can be found in Appendix 1.

The data is collected under a different setting compared to the human experiments reported in the previous deep imitation learning and RL literature [27, 39, 14]. Our experimental settings resulted significantly better human performance; see Appendix 1 for a human game score comparison. The high human game scores pave the way for a more insightful comparison between human and deep RL performances in terms of decision making. The dataset is available upon request to encourage future research in vision science, visuomotor behaviors, imitation learning, and reinforcement learning.

## 4  Gaze Network

Computer vision research has formalized visual attention modeling as an end-to-end saliency prediction problem, whereby a deep network can be used to predict a probability distribution of the gaze (saliency map). The ground truth saliency map is generated by converting discrete gaze positions to a continuous distribution using a Gaussian kernel with $\sigma$ equals to 1 visual degree, as suggested by [3].

We use the same deep neural network architecture and hyperparameters for all eight games (this is true for all models used in this work). The network architecture we use (shown in Fig. 2) is a three-channel convolution-deconvolution network. The inputs to the top channel are the images where the preprocessing procedure follows [27] and hence consist of a sequence of 4 frames stacked together where each frame is $84 \times 84$ in grayscale. The reason to use 4 frames is because a single image state is non-Markovian in Atari games, e.g., the direction of a flying bullet is ambiguous if we only see a single frame. The mid channel models motion information (optical flow) which is included since human gaze is sensitive to movement, and motion information has been used to improve gaze prediction accuracy [25]. Optical flow vectors of two continuous frames are calculated using the algorithm in [9] and fed into the network. The bottom channel includes bottom-up saliency map computed by the classic Itti-Koch model [16]. The output of the network is a gaze saliency map trained with Kullback-Leibler divergence as the loss function:

$$KL(P,Q) = \sum_i Q_i \log\left(\epsilon + \frac{Q_i}{\epsilon + P_i}\right) \quad (1)$$

where P denotes the predicted saliency map and Q denotes the ground truth. The regularization constant $\epsilon$ is set to $1e-10$.

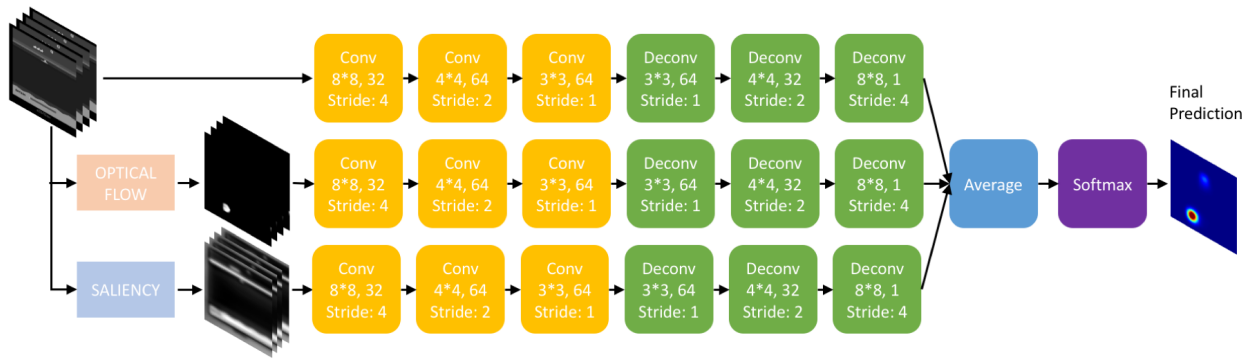

Figure 2: The three-channel gaze network. The top channel takes in images, the mid channel takes in the corresponding optical flow, and the bottom channel takes in the bottom-up image saliency. We then average the output of three channels. The final output is a gaze saliency map that indicates the predicted probability distribution of the gaze. The design of the convolutional layers follows the Deep Q-network [27].

For a performance comparison we use the classic bottom-up saliency model [16] as the first baseline (Saliency(S) in Table 1). Then we compute optical flow (Motion(M) in Table 1) of the current image as the second baseline, since motion is a reasonable indicator of visual attention. Then an ablation study is performed where the model consists only one or two channels of the original network in Fig. 2, i.e., Image(I), Image+Saliency(I+S), or Image+Motion(I+M). The performance of the algorithms are evaluated using four standard metrics in the visual saliency literature [33]: Normalized Scanpath Saliency (NSS), Area Under the Curve (AUC), Kullback-Leibler divergence (KL), and Correlation Coefficient (CC).

The quantitative results are shown in Table 1. Overall, the prediction results of our models are highly accurate across all games and largely outperform the saliency and motion baselines, indicated by the high AUC (above 0.93 for all games) obtained. A two-channel model (Image+Motion) in general achieves the best results. Further removing the motion information (having only the image) results in only slightly less accuracy–with the exception of the game Venture in which the speed of the monsters matters the most, hence removing motion decreases prediction accuracy. Including bottom-up saliency into the model does not improve the performance overall. This indicates that in the given tasks, the top-down visual attention is different than and hard to be inferred from the traditional bottom-up image saliency.

We encourage readers to view the video demo of the prediction results at `https://www.youtube.com/watch?v=-zTX9VFSFME`. Example predictions for all games are shown in Fig. 3 where the predicted gaze saliency map and ground truth human gaze positions are overlayed on top of the game frames. It is worth noticing that the prediction could be multimodal as in Figs 3d and 3g, indicating the task requires divided attention in these situations which our model successfully captured.

|  |  | Break-out | Free-way | Enduro | River-raid | Sea-quest | Ms-Pacman | Centi-pede | Ven-ture |
|---|---|---|---|---|---|---|---|---|---|
| Saliency(S) | NSS↑ | -0.075 | -0.175 | -0.261 | 0.094 | -0.208 | -0.376 | 0.665 | 0.422 |
| Motion(M) |  | 2.306 | 1.015 | 0.601 | 1.200 | 2.016 | 0.891 | 1.229 | 1.004 |
| Image(I) |  | 6.336 | 6.762 | 8.455 | 5.776 | 6.417 | 4.522 | 5.147 | 5.429 |
| I+S |  | 6.363 | 6.837 | 8.379 | 5.746 | 6.384 | 4.518 | 5.215 | 5.469 |
| I+M |  | **6.432** | **6.874** | **8.481** | 5.834 | 6.485 | **4.600** | **5.445** | **6.222** |
| I+S+M |  | 6.429 | 6.852 | 8.435 | **5.873** | **6.510** | 4.571 | 5.369 | 6.125 |
| Saliency(S) | AUC↑ | 0.494 | 0.560 | 0.447 | 0.494 | 0.352 | 0.426 | 0.691 | 0.607 |
| Motion(M) |  | 0.664 | 0.697 | 0.742 | 0.738 | 0.779 | 0.664 | 0.729 | 0.643 |
| Image(I) |  | **0.970** | **0.973** | **0.988** | **0.962** | 0.963 | 0.932 | 0.956 | 0.957 |
| I+S |  | 0.969 | **0.973** | **0.988** | 0.961 | 0.963 | 0.933 | 0.957 | 0.956 |
| I+M |  | **0.970** | 0.972 | **0.988** | **0.962** | 0.964 | 0.935 | **0.961** | **0.964** |
| I+S+M |  | 0.969 | **0.973** | **0.988** | **0.962** | 0.964 | **0.936** | 0.960 | **0.964** |
| Saliency(S) | KL↓ | 4.375 | 4.289 | 4.517 | 4.235 | 4.744 | 4.680 | 3.774 | 3.868 |
| Motion(M) |  | 13.097 | 10.638 | 8.312 | 9.151 | 9.133 | 12.173 | 10.810 | 12.853 |
| Image(I) |  | 1.304 | 1.261 | 0.834 | 1.609 | 1.464 | 1.985 | 1.711 | 1.749 |
| I+S |  | 1.301 | 1.260 | 0.834 | 1.613 | 1.470 | 1.995 | 1.709 | 1.727 |
| I+M |  | **1.294** | **1.257** | **0.832** | 1.593 | 1.438 | **1.959** | **1.622** | 1.512 |
| I+S+M |  | 1.299 | 1.260 | 0.835 | **1.592** | **1.437** | 1.961 | 1.645 | **1.510** |
| Saliency(S) | CC↑ | -0.009 | -0.023 | -0.033 | -0.008 | -0.035 | -0.048 | 0.065 | 0.048 |
| Motion(M) |  | 0.205 | 0.099 | 0.077 | 0.125 | 0.190 | 0.092 | 0.132 | 0.105 |
| Image(I) |  | 0.583 | 0.588 | 0.705 | 0.505 | 0.558 | 0.439 | 0.481 | 0.483 |
| I+S |  | 0.583 | 0.588 | 0.702 | 0.503 | 0.555 | 0.436 | 0.479 | 0.488 |
| I+M |  | **0.584** | **0.591** | **0.706** | 0.509 | **0.564** | **0.441** | **0.499** | **0.543** |
| I+S+M |  | **0.584** | 0.589 | 0.704 | **0.511** | 0.562 | 0.440 | 0.492 | 0.541 |

Table 1: Quantitative results of predicting human gaze across eight games. Random prediction baseline: NSS = 0.000, AUC = 0.500, KL = 6.159, CC = 0.000. For comparison, the classic [16] model (Saliency) and optical flow (Motion) are compared to versions of our model. All our models are accurate in predicting human gaze (AUC>0.93). In general the Image+Motion (I+M) model achieves the best prediction accuracy across games and four metrics.

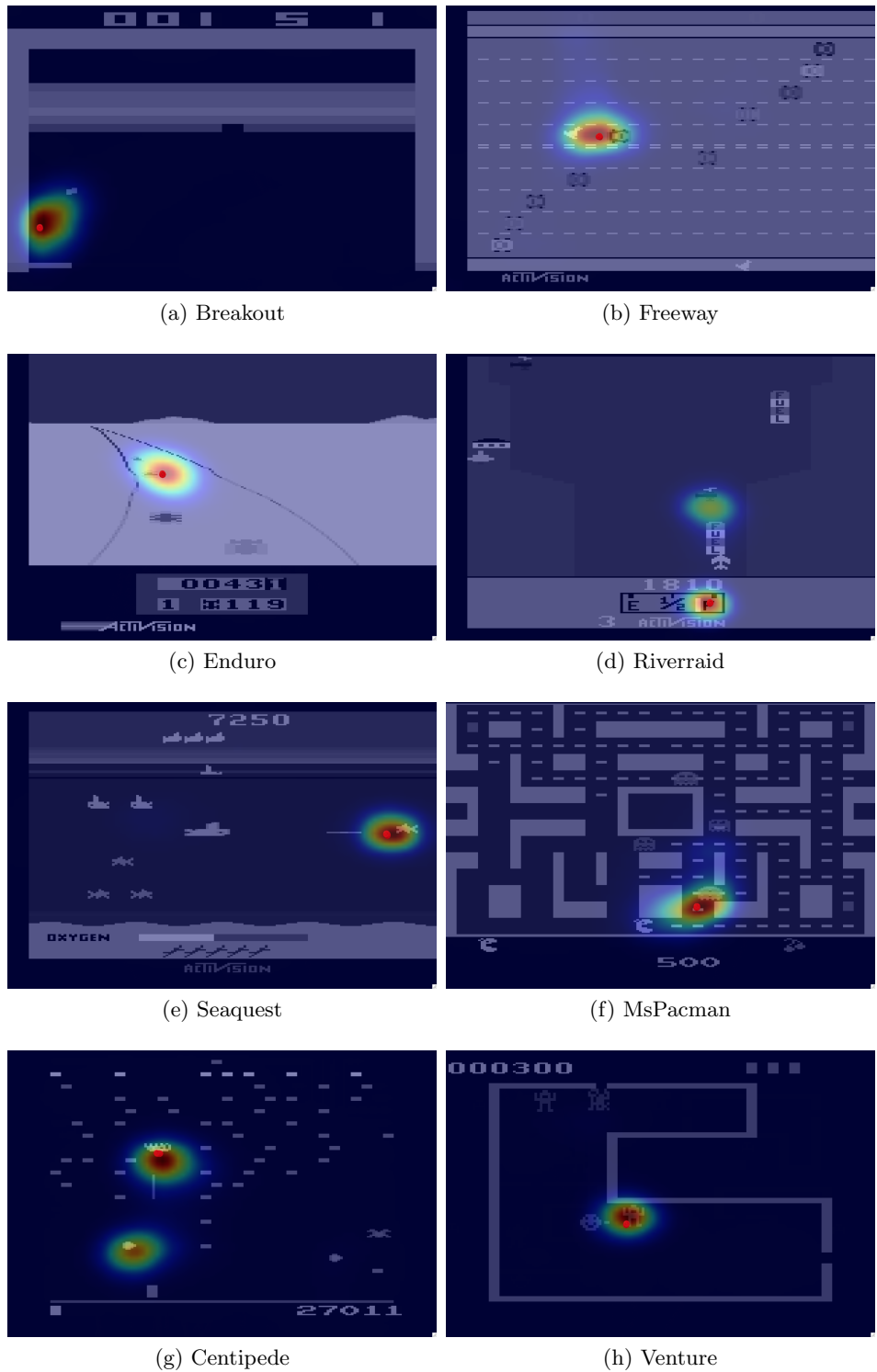

Figure 3: Visualization of gaze prediction results for eight games (best viewed in color). The solid red dot indicates the ground truth human gaze position. The heatmap shows the model's prediction as a saliency map, computed using the Image+Motion gaze network.

#### 4.0.1 Sample efficiency

In imitation learning tasks, sample efficiency is a major concern since human demonstration data could be expensive to collect. The proposed AGIL framework cannot be claimed to have an advantage over previous imitation learning or RL approaches if learning the attention model requires significantly more data. We study the effect of varying training sample size on prediction accuracy and find that the Image+Motion model is able to achieve high AUC values (above 0.88 for MsPacman and above 0.94 for 7 other games) with a single trial of human gaze data (15-minute)–although additional data can still help. The learning curves plotted against training sample size for all games can be found in Appendix 2. Therefore, training the gaze network does not incur a burden on sample size for the given task.

#### 4.0.2 Generalization across subjects

Do human subjects exhibit different gaze behaviors when performing the same task? This question is further investigated by training the gaze network on one subject's data and testing on the others for all games. We find that the gaze model is most accurate when trained and tested on the same subject. When tested on a different subject, the average prediction accuracy loss, in terms of correlation coefficient, is 0.091 comparing to trained and tested on the same subject (0.387 vs. 0.478). A detailed analysis can be found in Appendix 3.

## 5  AGIL: Policy Network with Attention

Given a gaze network that can accurately predict visual attention, we can incorporate it into a policy network that predicts a human's decisions. A deep network is trained with supervised learning to classify human actions given the current frame. The baseline network architecture follows the Deep Q-Network [27]. Here, we discuss two models that incorporate visual attention information to the imitation learning process.

### 5.1  Foveated rendering

One way to utilize gaze information is to reconstruct a biologically plausible representation of the visual stimulus (the stimulus as perceived by the human subject). We hypothesize that training the network with realistic retinal images may improve prediction, since these images are closer to the true human representation. We fed the visual angle of the game screen (44.58×28.50), a single ground truth human gaze position, and the original image into the Space Variant Imaging system [31]. The algorithm provides a biologically plausible simulation of foveated retinal images as shown in Fig. 1b by down-sampling and blurring the image according to the distribution of ganglion cells on the human retina.

The foveated images have a nice property of emphasizing the visual features near the gaze position. However, humans do not feel like they perceive the world like Figs. 1b, since memory plays an important role in reconstructing the visual world. A foveated image highlights the visual information being perceived at the moment, but it may lose other task-relevant information stored in memory. To compensate for this effect, we feed both the original image and the foveated image into a two-channel deep network. The model is referred as the Foveated model.

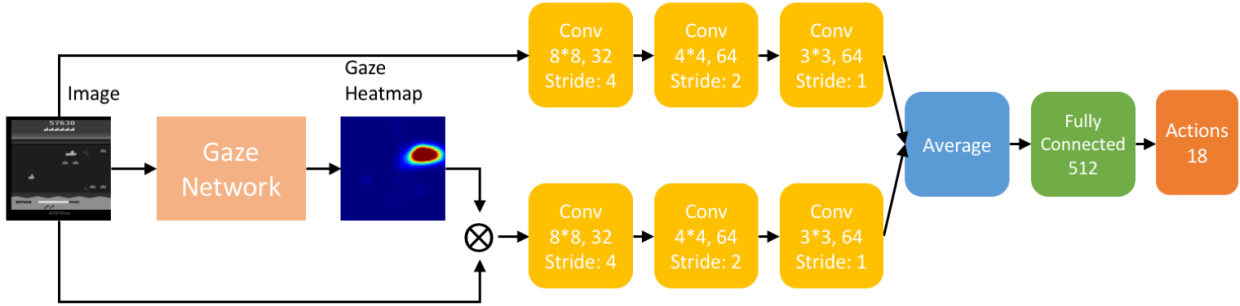

Figure 4: The policy network architecture for imitating human actions. The top channel takes in the current image frame and the bottom channel takes in the masked image which is an element-wise product of the original image and predicted gaze saliency map by the gaze network. We then average the output of the two channels.

The prediction results are shown in Table 2. As expected, the Foveated model consistently achieves better or comparable performance over the plain imitation model.

## 5.2 Masking with Attention

The foveated rendering approach directly incorporates human ground truth gaze into imitation learning. However, we argue that using the gaze heatmap learned by the gaze network might be better for two reasons: 1) While the ground truth gaze position is a single location, the human attention may be distributed on multiple objects (e.g., in Figs. 3d, 3e, and 3g); 2) The ground truth human gaze could be noisy but the predicted attention is accurate and clean. In addition, when the agent actually plays the game the ground truth human gaze will not be available.

We treat the predicted gaze heatmap as a saliency mask and multiply the mask with image frame element-wise. Similar to foveated rendering, the mask has the effect of emphasizing the stimulus being attended. For the same reason as in the Foveated model, we add a second channel that also takes the original image as the input and call this the Attention model. The final architecture is shown in Fig. 4. The prediction results are shown in Table 2. It is evident that the incorporating attention model has an advantage over the baseline. In particular, results for the four games that often require multitasking show large improvements: 15.6% on Seaquest, 16% on MsPacman, 5.1% on Centipede, and 6.6% on Venture. We conclude that including gaze information–either by foveated rendering or masking–can significantly improves the performance in terms of policy matching accuracy.

## 6 Evaluating the Learned Policy

The behavior matching accuracy is not the sole performance evaluation metric, since the ultimate goal of imitation learning is to learn a good policy to actually perform the task. When playing the games, the AGIL framework takes game images as input to the trained gaze network, and passes gaze network's predicted attention mask to the policy network to make decisions. The agent chooses an action $a$ probabilistically using a softmax function with Gibbs (Boltzmann) distribution

|           | Imitation      | +Foveated      | +Attention     |
|-----------|----------------|----------------|----------------|
| Breakout  | 81.5 ± 0.3     | 84.2 ± 0.1     | **86.2 ± 0.2** |
| Freeway   | **96.7 ± 0.0** | 96.4 ± 0.1     | 96.4 ± 0.2     |
| Enduro    | 60.6 ± 0.4     | 60.5 ± 0.4     | **61.9 ± 0.3** |
| Riverraid | **72.5 ± 0.3** | 72.4 ± 0.4     | **72.5 ± 0.4** |
| Seaquest  | 46.0 ± 1.8     | 51.4 ± 1.0     | **61.6 ± 0.2** |
| MsPacman  | 54.6 ± 1.0     | 65.1 ± 1.0     | **70.6 ± 0.3** |
| Centipede | 61.9 ± 0.2     | 64.8 ± 0.3     | **67.0 ± 0.3** |
| Venture   | 46.7 ± 0.2     | 48.7 ± 0.1     | **53.3 ± 0.3** |

Table 2: Percentage accuracy (mean ± standard deviation) in predicting human actions across eight games using different models. Random prediction baseline: 5.56. The model in Fig. 4 (+Attention) yields the best prediction accuracy.

according to policy network's prediction $P(a)$:

$$\pi(a) = \frac{\exp(\eta P(a))}{\sum_{a' \in \mathcal{A}} \exp(\eta P(a'))} \qquad (2)$$

where $\mathcal{A}$ denotes the set of all possible actions, $\exp(.)$ denotes the exponential function, and the temperature parameter $\eta$ is set to 1.

The average games scores over 100 episodes per game are reported in Table 3, in which each episode is initialized using a randomly generated game seed to ensure enough variability in game dynamics [12]. Our model with attention outperforms the previous plain imitation learning results of [14] and the one without attention using our dataset. The improvement over the latter is 3.4% to 1143.8%. The improvement is minor for Freeway since the scores are close to the maximum possible score (34.0) for this task.

An advantage of imitation learning comparing to RL is its sample efficiency. We show performance of deep Q-learning [27] implemented using the standard OpenAI DQN benchmark [6] trained for the same sample size (training sample size for each game can be found in Appendix 1). It is evident that DQN's performance is not at the same level with imitation learning given the same amount of training data. Remarkably, after 200 million training samples (corresponds to about 38.58 days of play experience when playing at 60Hz), our method is still better than or comparable to DQN in four games: Freeway, Enduro, Centipede, and Venture. In fact, our method achieves the state-of-the-art result on Centipede comparing to any RL methods or their combinations [13].

Why does the learned visual attention model improve action prediction accuracy and task performance? First, attention highlights task-relevant visual features in a high-dimensional state space, even though the features may only occupy a few pixels in that space, as observed in Figs 3. Hence, attention can be seen as a feature selection mechanism that biases the policy network to focus on the selected features. Second, attention could help to identify and disambiguate the goal of current action when multiple task-relevant objects are present. For example, in Fig. 5b and 5c the gaze indicates that the goal of current action involves the enemy to the left or above the yellow submarine. The corresponding actions would be moving left for the first case and moving up for the second. The two enemies are visually identical hence the learning agent cannot predict the correct action without gaze information – an issue further exacerbated by convolutional layers of the network due to their spatial invariant nature. For these reasons, modeling human attention help the agent infer the correct decision state of the human teacher and understand the underlying reason for that decision.

|          | Imitation [14] | Imitation Our data   | AGIL Our data          | Improvement | DQN Same $N$ | DQN $N=200M$ |
|---------:|:--------------:|:--------------------:|:----------------------:|:-----------:|:------------:|:------------:|
| Breakout | 3.5            | $1.6 \pm 1.2$        | $\mathbf{19.9 \pm 14.1}$      | 1143.8%     | 1.52         | 401.2        |
| Freeway  | 22.7           | $29.6 \pm 1.2$       | $\mathbf{30.6 \pm 1.2}$       | 3.4%        | 0            | 30.3         |
| Enduro   | 134.8          | $239.8 \pm 90.8$     | $\mathbf{295.7 \pm 99.5}$     | 23.3%       | 0            | 301.8        |
| Riverraid| 2148.5         | $2419.7 \pm 655.8$   | $\mathbf{3338.5 \pm 1485.9}$  | 38.0%       | 1510         | 8316         |
| Seaquest | 195.6          | $252.2 \pm 109.2$    | $\mathbf{788.9 \pm 609.2}$    | 212.8%      | 100          | 5286         |
| MsPacman | 692.4          | $1069.9 \pm 810.5$   | $\mathbf{1755.1 \pm 1000.9}$  | 64.0%       | 230          | 2311         |
| Centipede| N/A            | $5543.0 \pm 3509.5$  | $\mathbf{9515.4 \pm 5626.8}$  | 71.7%       | 2080         | 8309         |
| Venture  | N/A            | $363.0 \pm 133.2$    | $\mathbf{468.0 \pm 176.6}$    | 28.9%       | 0            | 380.0        |

Table 3: A comparison of game scores (mean ± standard deviation) between plain imitation learning from a previous work [14], plain imitation learning using our dataset, AGIL, and deep reinforcement learning (DQN) [27, 6]. The DQN scores are recorded at two different training sample sizes: one at the same sample size with our dataset (114K-223K depends on the game) and the other one at 200 million samples. The "Improvement" column indicates AGIL's performance increase over the plain imitation learning using our dataset.

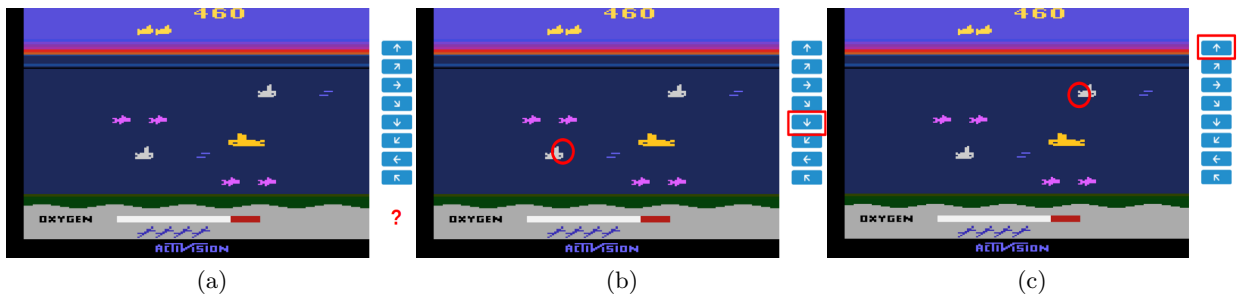

Figure 5: Human gaze information helps the learning agent correctly infer the underlying reason for the chosen action. The red circles indicate human teacher's gaze position.

# 7 Conclusion and Future Work

A research question for imitation learning in visuomotor tasks is: what should be learned from the human teacher? The agent could learn the policy (behavior cloning), the reward function (inverse RL), or some high-level cognitive functionality such as visual attention. To our knowledge, the proposed AGIL framework represents the first attempt to learn visual attention for imitation learning tasks. Through modeling the human teacher's visual attention the student agent gains a deeper understanding of *why* a particular decision is made by the teacher. We show that it is feasible to learn an accurate task-driven human visual attention model, and that combining this with deep imitation learning yields promising results.

The high accuracy achieved in predicting gaze in our work implies that, given a cognitively demanding visuomotor task, human gaze can be modeled accurately using an end-to-end learning algorithm. This suggests that popular deep saliency models could be used to learn visual attention, given task-driven data.

In this student-teacher paradigm, a better learning framework is possible when we have more knowledge on both the student and the teacher. There is much room for future work to understand more about the human teacher from a psychological perspective. Due to human visuomotor reaction time, action $a_t$ may not be conditioned on the image and gaze at time $t$, but on images and gazes several hundreds milliseconds prior. More importantly, the human memory system allows for states of previously attended objects to be preserved, and an internal model may perform model-based prediction to update the environmental states in memory. These cognitive functionalities could be readily implemented by deep networks models, such as a recurrent neural network to allow for a better prediction of human actions.

The results of [27] have demonstrated the effectiveness of end-to-end learning of visuomotor tasks, where the DQN excels at games that involve a single task. However, for games such as Seaquest and MsPacman–which typically involve multiple tasks–the performance is still below human levels, even though human reaction time was limited in their setting. In addition, DQN takes millions of samples to train. The above issues could be potentially alleviated by combining AGIL and deep RL where attention model can help extract features to speedup learning and to indicate task priority. By making our dataset publicly available we encourage future research towards the combined approach.

## Appendix 1: Dataset

Our goal is to obtain the best possible control policy from human subjects, hence we take into account the limitations of human reaction time and fatigue. Our experimental setting resulted in significantly better human performance than those reported in previous literature, as seen in Table 4. The key for improvement is to allow human players to have enough decision time by implementing the step-wise play mode described in the paper. The statistics of our dataset is also reported.

## Appendix 2: Gaze Network Learning Curve

Fig. 6 shows the learning curve of the best performing Image+Motion model, where AUC on the testing dataset is plotted against the training data size. The main observation is that training the gaze network for AGIL is sample efficient.

## Appendix 3: Generalization Across Human Subjects

To test whether gaze behavior generalizes across subjects, we show correlation coefficients for the gaze network that is trained using one trial (15-minute) of the data from a single subject and tested on trials from other subjects (15-minute each). The results are shown in Fig. 7. In general, the gaze model is more accurate when train and test on the same subjects, indicated by higher correlation coefficient values on the diagonal. The prediction accuracy decreases when training and testing on different subjects by as much as 0.419 (Breakout, trained on Subject3 and tested on Subject1). When tested on a different subject, the average prediction accuracy loss, in terms of correlation coefficient, is 0.091 comparing to trained and tested on the same subject (0.387 vs. 0.478). Meanwhile, the within-subject variance is considerably smaller. When trained on one subject's data and tested on the same subject, the average mean deviation across subjects and games is 0.011.

|          | Mnih  | Wang    | Hester  | Our Data | #Trials     | #Training | #Testing |
|----------|-------|---------|---------|----------|-------------|-----------|----------|
|          | [27]  | [39]    | [14]    |          | 15min/trial | (frames)  | (frames) |
| Breakout | 31.8  | 30.5    | 79      | 508      | 9           | 119,701   | 13,704   |
| Freeway  | 29.6  | 29.6    | 32      | 33       | 8           | 114,740   | 17,062   |
| Enduro   | 309.6 | 860.5   | 803     | 1,104    | 12          | 185,408   | 17,355   |
| Riverraid| 13,513| 17,118  | 39,710  | 52,690   | 12          | 181,049   | 16,460   |
| Seaquest | 20,182| 42,054.7| 101,120 | 201,130  | 12          | 180,630   | 17,015   |
| MsPacman | 15,693| 6,951.6 | 55,021  | 92,610   | 12          | 187,176   | 17,058   |
| Centipede| 11,963| 12,017  | N/A     | 152,967  | 12          | 167,587   | 15,714   |
| Venture  | 1,188 | 1,187.5 | N/A     | 11,800   | 12          | 223,102   | 23,621   |

Table 4: Left: A comparison of best human scores for eight Atari games across datasets. The experimental settings we have lead to better performance than the the results from [27], [39], and the best scores in [14]. Right: Statistics of our dataset.

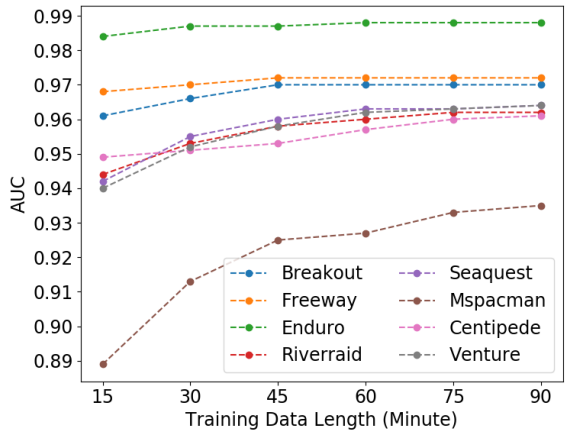

Figure 6: Learning curve for all eight games, i.e., the AUC value obtained when training on different amount of data and testing on the same holdout dataset. The model is able to achieve high AUC values with limited amount of data. Note that the we do not show the full scale of y-axis for better visualization.

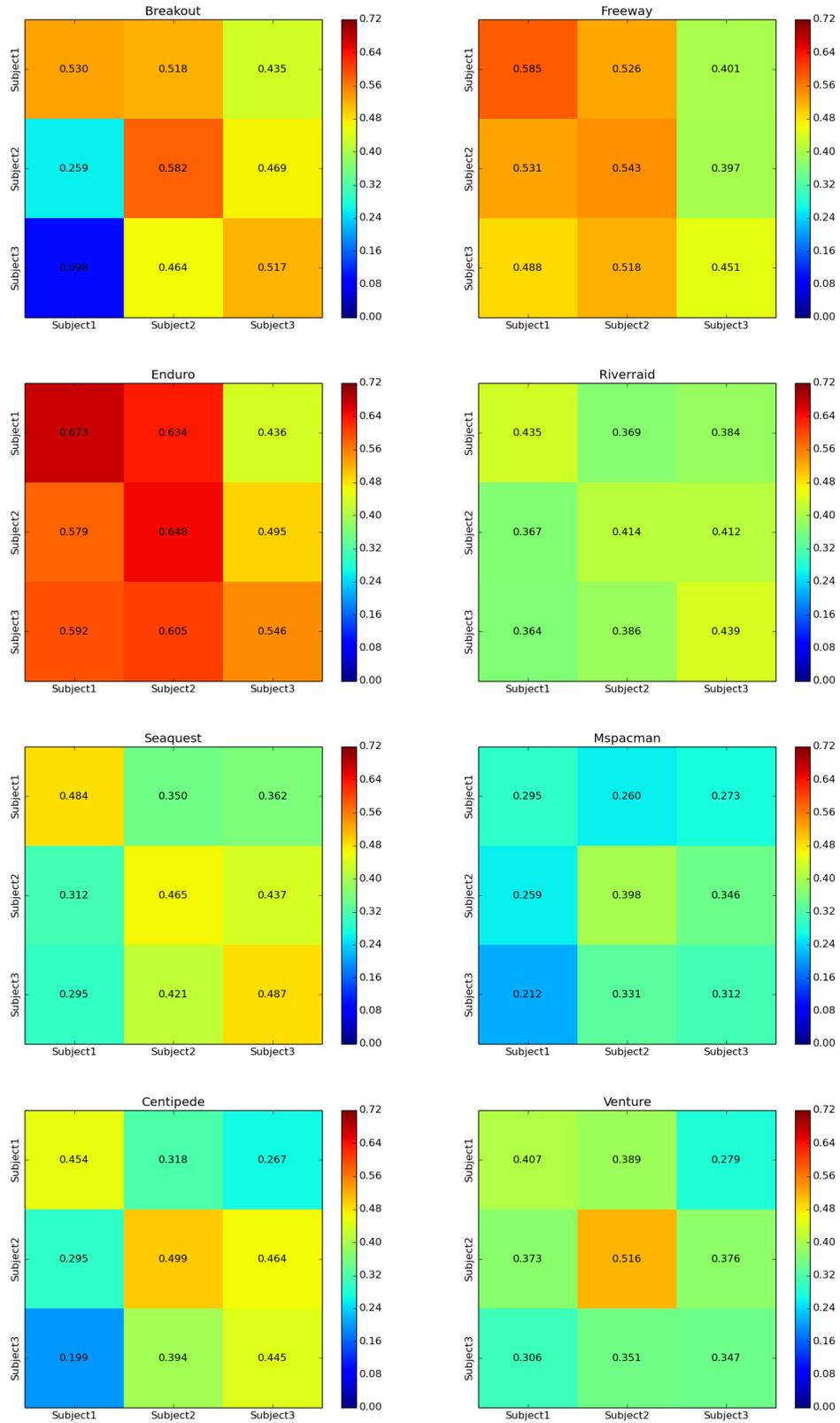

Figure 7: Correlation coefficient matrices of the gaze network when trained on one subject and tested on another subject. Y axis: Subject ID of the training data. X axis: Subject ID of the testing data.